# Shadow Image Enlargement Distortion Removal

Raid R. Al-Nima, Ali N. Hamoodi, Radhwan Y. Al-Jawadi, and Ziad S. Mohammad
*Northern Technical University / Mosul / Iraq*

*Abstract*—This project aims to adopt preprocessing operations to get less distortions for shadow image enlargement. The preprocessing operations consists of three main steps: first enlarge the original shadow image by using any kind of interpolation methods, second apply average filter to the enlargement image and finally apply the unsharp filter to the previous averaged image. These preprocessing operations leads to get an enlargement image very close to the original enlarge image for the same shadow image. Then comparisons established between the adopted image and original image by using different types of interpolation and different alfa values for unsharp filter to reach the best way which have less different errors between the two images.

## I. INTRODUCTION

Polynomial interpolation methods have been studied quite extensively in the signal and image processing literature of the past three decades[1-3].

Interpolation is the process of estimating the values of a continuous function from discrete samples. Image processing applications of interpolation include image magnification or reduction, subpixel image registration, to correct spatial distortions, and image decompression, as well as others of the many image interpolation techniques available, nearest neighbour, bilinear and cubic convolution are the most common, and will be talked about here. Since interpolation provides a perfect reconstruction of a continuous function, provided that the data was obtained by uniform sampling at or above the nyquist rate. Image data is generally acquired at a much lower sampling rate. The mapping between the unknown high-resolution image and the low-resolution image is not invertible. One of the essential aspects of interpolation is efficiency since the amount of data associated with digital images is large.

For the related works, An image adaptive threshold imposed on the multi scale products is calculated to identify the significant structures in 2003 [4]. Then, a camera noise model are described and show how it can be combined with a set of parameterized camera response functions to develop a prior on the noise response by estimate the noise level function in 2006 [5]. Also, a method for automatic noise estimation and removal of noise is proposed but from a single image in 2008 [6]. Moreover, for shadow image (medical X-Ray) a noise removal method is proposed by wavelet domain in 2008 [7].

This study aims to adopt preprocessing operations to get less distortions for shadow image enlargement (i.e get an enlargement shadow image very close to the original enlarge image).

## II. IMAGE ZOOMING METHODS

The general form for an interpolation function is:
$$g(x) = \sum_K c_K\, u(distance_K) \quad \ldots\ldots\ldots\ldots\ldots\ldots\ldots..(1)$$
where g( ) is the interpolation function, u( ) is the interpolation Kernel, $distance_K$ is the distance from the point under consideration, x, to the grid point, $x_K$ and $c_K$ are the interpolation coefficients. The $c_K$'s are chosen such that $g(x_K) = f(x_K)$ for all $x_K$. this means that the grid point values should not change in the interpolation image.

### A. Nearest neighbour interpolation

Nearest neighbour interpolation, the simplest method, determines the grey level value from the closest pixel to the special input coordinates, and assigns that value to the output coordinates. It should be noted that this method does not really interpolate values, it just copies existing values[4]. Since it does not alter values, it is preferred if subtle variations in the grey level values need to be retained.

For one-dimension nearest neighbour interpolation, the number of grid points needed to evaluate the nearest neighbour interpolation, the number of grid points needed to evaluate the interpolation function is four.

For nearest nighbour interpolation, the interpolation kernel for each direction is:

$$u(s) = \begin{cases} 0 & |s| > 0.5 \\ 1 & |s| < 0.5 \end{cases} \quad \ldots\ldots\ldots\ldots\ldots\ldots\ldots.(2)$$

where s is the distance between the point to be interpolated and the grid point being considered. The interpolation coefficients $C_K = f(x_K)$.

### B. Bilinear interpolation

Bilinear interpolation determine the grey level value from the weighted average of the four closest pixels to the specified input coordinates, and assigns that value to the output coordinates[8].

First, two linear interpolations are performed in one direction (horizontally in this paper) and then one more linear interpolation is performed in the perpendicular direction (vertically in this paper).

For one-dimension linear interpolation, the number of grid points needed to evaluate the interpolation function is two. For bilinear interpolation (linear interpolation in two dimensions), the number of grid points needed to evaluate the interpolation function is four.

For linear interpolation, the interpolation kernel is:

$$u(s) = \begin{cases} 0 & |s| > 1 \\ 1-|s| & |s| < 1 \end{cases} \quad \ldots\ldots\ldots\ldots\ldots(3)$$

where s is the distance between the point to be interpolated and the grid point being considered. The interpolation coefficients $C_K = f(x_K)$.

## C. Cubic convolution interpolation

Cubic convolution interpolation determines the grey level value from the weighted average of the 16 closest pixels to the specified input coordinates, and assigns that value to the output coordinates [8].

The image is slightly sharper than produced by bilinear interpolation, and it does not have the disjointed appearance produced by nearest neighbour interpolation.

For one-dimension cubic convolution interpolation, the number of grid points needed to evaluate the interpolation function is four, two grid points on either side of the point under consideration. For bicubic interpolation (cubic convolution interpolation in two dimensions), the number of grid points needed to evaluate the interpolation function is 16, two grid points on either side of the point under consideration for both horizontal and vertical directions. The grid points needed in one dimention and two dimension cubic convolution interpolation are shown in Fig. 1.

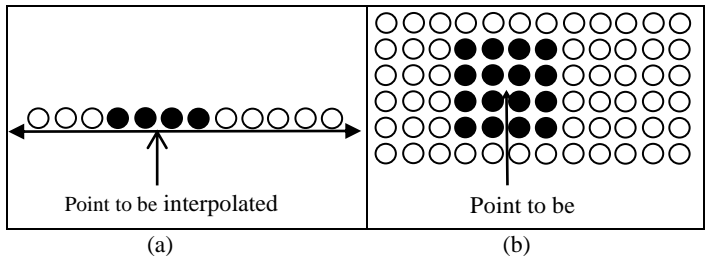

(a)           (b)
Fig. 1: (a) Grid points need in one-dimension and (b) Grid points need in two-dimension convolution interpolation

The one-dimension cubic convolution interpolation kernel is[9-10]:

$$u(s) = \begin{cases} 3/2|s|^3 - 5/2|s|^2 + 1 & 0 \le |s| < 1 \\ -1/2|s|^3 - 5/2|s|^2 - 4|s| + 2 & 1 \le |s| < 2 \\ 0 & 2 < |s| \end{cases} \quad \ldots\ldots\ldots\ldots(4)$$

Where s is the distance between the point to be interpolated and the grid point being considered. A plot of the one-dimension cubic convolution interpolation kernel $v_s$ s is shown in Fig. 2:

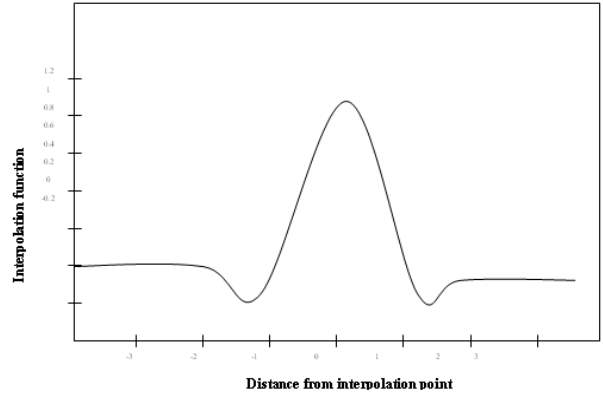

Fig.2: Cubic convolution interpolation kernel $v_s$ s

For two dimensional interpolation, the one-dimensional interpolation function is applied in both directions. It is a separable extension of the one-dimensional interpolation function. Given a point (x,y) to interpolate, where $x_K < x < x_{K+1}$ and $y_K < y < y_{K+1}$, the two-dimensional cubic convolution interpolation function is:

$$g(x,y) = \sum_{l=-1}^{2}\sum_{m=-1}^{2} C_{j+l,K+m}\, u(dis\tan ce_x)\, u(dis\tan ce_y) \quad \ldots\ldots\ldots\ldots..(5)$$

where u( ) is the interpolation function of equation (4), and $distance_x$ and $distance_y$ are the x and y distances from the four grid points in each direction. For non-boundary points, the interpolation coefficients, $C_{jK}$'s are given by $C_{jK} = f(x_j, y_K)$.

## D. Computational and Computation time

Nearest neighbour interpolation is the most efficient in terms of computation time. Bilinear interpolation requires 2 to 4 times the computation time of nearest neighbour interpolation. Cubic convolution interpolation requires about 10 times the computation time of nearest neighbour interpolation.

## E. Accuracy

Nearest neighbour interpolation generally performs poorly. This image may be spatially offset by up to ½ a pixel, causing a jagged or blocky appearance. Bilinear interpolation generates and image of smoother appearance than nearest neighbour interpolation, but the grey levels are altered in the process, resulting in blurring or loss of image resolution.

## III. THE PREPROCESSING OPERATIONS

The preprocessing operations consists of three main steps: first enlarge the original shadow image (See Fig. 3) by using any kind of the previous interpolated methods, second apply average filter to the enlargement image and finally apply the unsharp filter to the previous averaged image.

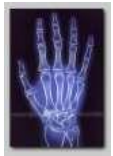

Fig. 3: The original shadow image before enlargement

### A. Average filter

An averaging filter is useful for removing grain noise from a photograph. Because each pixel gets set to the average of the pixels in its neighborhood, local variations caused by grain are reduced. That each output pixel is set to an "average" of the pixel values in the neighbourhood of the corresponding input pixel [8]. Thus, the coefficients of this filter are:

| $P_1$ | $P_2$ | $P_3$ |
|---|---|---|
| $P_4$ | $P_{AV}$ | $P_5$ |
| $P_6$ | $P_7$ | $P_8$ |

Where:

$P_1, P_2, \ldots P_8$: The neighbourhood pixels.
$P_{AV}$: The middle pixel of mask.

So, the middle coefficient is calculated as average in the output image by equation (2.10):

$$P_{AV} = \frac{(P_1 + P_2 + \ldots\ldots + P_8)}{8} \quad \ldots\ldots\ldots\ldots\ldots\ldots(6)$$

### B. Unsharp filter

The unsharp masking enhancement algorithm is a representative example of the practical image sharpening methods. Image sharpening deals with enhancing detail information in an image. The detailed information is typically contained in the high spatial frequency components of the image. The details of information includes edges and, in general, corresponds to image features that are spatially small. This piece of information is visually important because it delineates object and feature boundaries and is important for textures in objects [11].

An unsharp masking filter which is created from the negative of the Laplacian filter with parameter alpha is shown below in equation (2.9):

$$-\nabla^2 \approx \frac{1}{(\alpha+1)} \begin{bmatrix} -\alpha & \alpha-1 & -\alpha \\ \alpha-1 & \alpha+5 & \alpha-1 \\ -\alpha & \alpha-1 & -\alpha \end{bmatrix} \ldots\ldots\ldots\ldots\ldots\ldots(7)$$

Alpha must be in the range of 0.0 to 1.0[8].

## IV. RESULTS

i- A comparison of the implementation of nearest neighbour interpolation with Matlab's implement-tation shows some interesting results. The interpolated images are shown in Fig. 4.

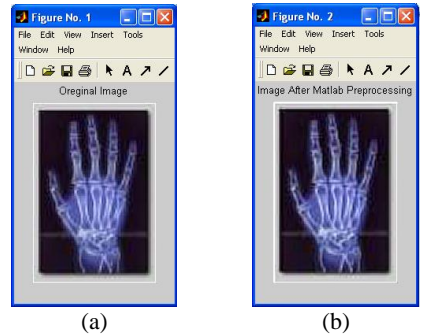

(a)      (b)

Fig. 4: (a) Original enlarge image and
(b) Matlab's implementation of nearest neighbour interpolation

ii- A comparison of the implementation of bilinear interpolation with Matlab's implementation shows some interesting results. The interpolated images are shown in Fig. 5.

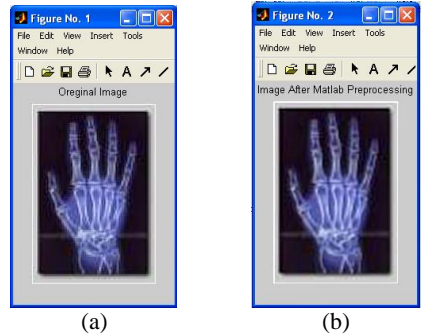

(a)      (b)

Fig. 5: (a) Original enlarge image and
(b) Matlab's implementation of bilinear interpolation

iii- A comparison of the implementation of bicubic interpolation with Matlab's implementation shows some interesting results. The interpolated images are shown in Fig. 6.

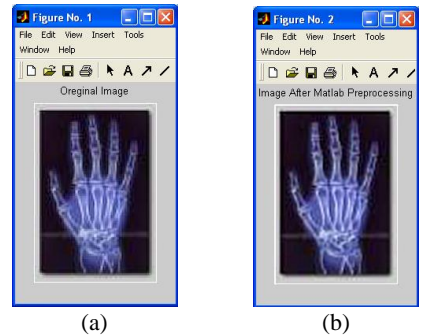

(a)      (b)

Fig. 6: (a) Original enlarge image and
(b) Matlab's implementation of bicubic interpolation

The error ratio (difference) between the original image and Matlab's implementation of nearest neighbour interpolation for different alfa are computation from program. The relationship between alfa and the error ratio (difference) is illustrated in Table I.

Table I

| Alfa | Error |
|---|---|
| 0.0 | 0.048300% |
| 0.1 | 0.048258% |
| 0.2 | 0.047799% |
| 0.3 | 0.047928% |
| 0.4 | 0.047707% |
| 0.5 | 0.047661% |
| 0.6 | 0.047447% |
| 0.7 | 0.047583% |
| 0.8 | 0.047560% |
| 0.9 | 0.047561% |
| 1.0 | 0.046225% |

The error ratio (difference) between the original image and Matlab's implementation of bilinear convolution interpolation for different alfa are computation from program. The relationship between alfa and the error ratio (difference) is illustrated in Table II.

Table II

| Alfa | Error |
|---|---|
| 0.0 | 0.045246% |
| 0.1 | 0.045245% |
| 0.2 | 0.044921% |
| 0.3 | 0.045062% |
| 0.4 | 0.044859% |
| 0.5 | 0.044853% |
| 0.6 | 0.044729% |
| 0.7 | 0.044823% |
| 0.8 | 0.044824% |
| 0.9 | 0.044824% |
| 1.0 | 0.043541% |

The error ratio (difference) between the original image and Matlab's implementation of bicubic convolution interpolation for different alfa are computation from program. The relationship between alfa and the error ratio (difference) is illustrated in Table III.

Table III

| Alfa | Error |
|---|---|
| 0.0 | 0.047914% |
| 0.1 | 0.047903% |
| 0.2 | 0.047507% |
| 0.3 | 0.047637% |
| 0.4 | 0.047402% |
| 0.5 | 0.047382% |
| 0.6 | 0.047237% |
| 0.7 | 0.047322% |
| 0.8 | 0.047315% |
| 0.9 | 0.047315% |
| 1.0 | 0.045980% |

The suggested method can be considered as one of suggestable techniques as in [12-70].

V. CONCLUSION

- Preprocessing operations can be implement to enlarge shadow image with less distortions (i.e very close to the original enlarge image).
- In the unsharpe filter, when the alfa value is big it gives best results than when it is small in the shadow image.
- The bilinear method of interpolation attained to the best results with alfa value equal to one.
- The adopted method success in shadow image enlargement.